# STCNet: Spatio-Temporal Cross Network for Industrial Smoke Detection

Yichao Cao, *Student Member, IEEE*, Qingfei Tang, Xiaobo Lu, Fan Li, and Jinde Cao, *Fellow, IEEE*

*Abstract*—**Industrial smoke emissions present a serious threat to natural ecosystems and human health. Prior works have shown that using computer vision techniques to identify smoke is a low cost and convenient method. However, industrial smoke detection is a challenging task because industrial emission particles are often decay rapidly outside the stacks or facilities and steam is very similar to smoke. To overcome these problems, a novel Spatio-Temporal Cross Network (STCNet) is proposed to recognize industrial smoke emissions. The proposed STCNet involves a spatial pathway to extract texture features and a temporal pathway to capture smoke motion information. We assume that spatial and temporal pathway could guide each other. For example, the spatial path can easily recognize the obvious interference such as trees and buildings, and the temporal path can highlight the obscure traces of smoke movement. If the two pathways could guide each other, it will be helpful for the smoke detection performance. In addition, we design an efficient and concise spatio-temporal dual pyramid architecture to ensure better fusion of multi-scale spatiotemporal information. Finally, extensive experiments on public dataset show that our STCNet achieves clear improvements on the challenging RISE industrial smoke detection dataset against the best competitors by 6.2%. The code will be available at: https://github.com/Caoyichao/STCNet.**

## I. INTRODUCTION

Industrial smoke emissions may cause adverse effects on human health and ecological environment. Large amounts of air pollutants may cause or contribute to an increase in mortality or serious illness or may pose a present or potential hazard to human health. Smoke detection technology based on computer vision can help regulators to obtain visual evidence and enterprises to implement self-monitoring. In addition, smoke is the main manifestation of early fires. Smoke detection is an efficient approach for large range fire monitoring.

In industrial smoke detection task, plants usually not only emit smoke, but also emit a lot of steam. Steam and smoke have very similar appearance, which brings a great challenge to smoke detection in this scene. Different from some smoke datasets (in which steam and smoke are not deliberately distinguished), RISE dataset [29] makes a clear distinction between steam and smoke.

More realistic industrial smoke emission dataset makes practical smoke detection method possible.

So far, there are a lot of references about the recognition of specific features of smoke. According to the dimension difference of input data, existing methods can be divided into image-based and video-based. Image-based smoke detection methods tend to detect smoke areas from single-frame image. Video-based methods usually not only learn spatial features from single frame, but also learn temporal information in temporal domain from videos.

In some cases, image-based method is a good choice when stable and reliable image sequences are not available. Tian *et al.* [2] proposed to detect and separate smoke from a single image frame by convex optimization. Yuan *et al.* [37] proposed to combine local binary pattern (LBP) like features, kernel principal component analysis (KPCA), and Gaussian process regression (GPR) for smoke detection. There are also some researches that apply convolutional neural networks (CNNs) to smoke detection and recognition. Yin *et al.* [38] proposed a deep normalization and convolutional neural network (DNCNN) with 14 layers for smoke recognition, in which batch normalization is used to speed up the training process and boost accuracy of smoke recognition.

However, the dynamic characteristics of smoke often play an important role in the recognition process. When the human eyes distinguish smoke from the video, dynamic features are often used as key reference information. If the recognition model could learn context information from sequence data, its recognition accuracy will be improved theoretically. From motion point of view, namely higher order linear dynamical system (h-LDS) descriptor was proposed as a dynamic texture descriptor for video smoke identification [5]. There are some researchers trying to use deep learning method for smoke detection. Lin *et al.* [6] proposed a joint detection framework for video smoke detection, in which faster RCNN is employed to generate suspected smoke boxes and 3D CNN is used to extract spatio-temporal features of the clip. However, this method also have weak points, such as the considerable computational complexity.

Despite these efforts, video-based smoke recognition is still a challenging task. Early smoke objects are usually small in size and the variance of smoke color, texture, and interference is large,

This work was supported by the National Natural Science Foundation of China (No.61871123), Key Research and Development Program in Jiangsu Province (No.BE2016739) and a Project Funded by the Priority Academic Program Development of Jiangsu Higher Education Institutions. We thank the Big Data Center of Southeast University for providing the facility support on the numerical calculations in this paper.

Yichao Cao, Xiaobo Lu and Jinde Cao is with School of Automation, Southeast University, Nanjing 210096, China.
Qingfei Tang was with Northeastern University, Shenyang, China. He is now with the Nanjing Enbo Technology Co., Ltd., Nanjing, 210007, China .
Fan Li is with School of Information Science and Engineering, Southeast University, Nanjing 210096, China.

which make industrial smoke detection very difficult. The key to accurate smoke detection is the spatiotemporal feature learning ability.

Given the aforementioned concerns, we propose a novel industrial smoke detection framework, denoted as Spatio-Temporal Cross Network (STCNet). Inspired by the two-stream methods [39][21], this framework attempts to cross fuse multi-scale spatial and temporal features in forward of the smoke detection process. The main contributions of this paper are as follows:

- A novel video smoke detection architecture utilizing residual frames is proposed, which can effectively focus on subtle smoke objects.
- The Spatio-Temporal Cross Network (STCNet) to integrate the spatial feature learning approach with the temporal modeling through dual pyramid, achieve collaborative promotion of two-path networks.
- Experimental results on challenging RISE dataset demonstrate our proposed method achieves state-of-the-art results on smoke detection task.

The rest of this paper is organized as follows. Section II summarizes related works. The proposed architecture for smoke representation described in Section III. Detailed performance studies and analysis are conducted in Section IV. Finally, conclusions and discussions are drawn in Section V. Our model code will be released if the paper is accepted.

## II. RELATED WORK

In this section, some representative smoke detection methods are reviewed. Although few literatures research on video smoke detection, there is substantial literature on video understanding. It is a nature idea to use video understanding methods for smoke recognition. Therefore, we use a subsection to introduce and discuss video understanding methods in details.

### A. Smoke Detection

The success of existing approaches for smoke detection rely on robust smoke feature description. Both smoke feature descriptors and deep learning methods can be divided into two categories according to the dimension of smoke features: image-based methods [2], [3], [4], [17], [18], [19], [20] and video-based methods [5], [6]. To motivate the rationale for the proposed methods, some representative works are reviewed from three aspects: image-based, video-based and deep learning-based.

Image based methods usually focus on the smoke texture, color, shape and edge. Yuan [18] proposed a double mapping framework for smoke detection, in which the first mapping calculates histograms of edge orientation, edge magnitude and Local Binary Pattern (LBP) bit, and densities of edge magnitude, LBP bit, color intensity and saturation, the second mapping computes the statistical characteristics of mean, variance, skewness, kurtosis and Hu moments. Some researchers formulated the smoke detection task as sparse representation and convex optimization problem [2], [19], [20]. Tian *et al.* [2] proposed to separate quasi-smoke and quasi-background components by dual over-complete dictionaries, in which the respective sparse coefficients is concatenated for smoke detection. Based on the airlight albedo ambiguity model, Long *et al.* [17] proposed to detect the smoke and predict thickness distribution through transmission.

Although image-based methods have achieved impressive results, these methods do not meet the requirements of the practical applications, since they ignore the motion information of the smoke. The dynamic characteristics of smoke often play an important role in the recognition process of human vision. Dimitropoulos *et al.* [5] proposed a higher order linear dynamical system (h-LDS) descriptor for multidimensional dynamic smoke texture analysis.

In recent years, Deep learning methods has achieved competitive results on various tasks, such as visual recognition [40], speech recognition [41] and natural language processing [42]. Researchers designed a deep normalization and convolutional neural network (DNCNN) for smoke detection, which is a superior alternative for traditional hand-crafted methods [38]. Zhao *et al.* [43] demonstrated the effectiveness of using saliency detection and deep convolutional neural network in localization and recognition of wildfire in unmanned aerial vehicle (UAV) images. Given the smoke candidate patch, the dark channel was reported to has more elaborate information of the smoke and the detailed features of dark channel images have been used as cue to perform smoke detection [4]. One of the difficulties in smoke recognition is the limitation of smoke sample number for training. To ease this limitation, Xu *et al.* [44] proposed a framework based fast detector SSD and MSCNN for smoke detection using synthetic smoke image samples.

Recently, there are some video-based smoke detection methods using deep learning [5]. Lin *et al.* [6] proposed a joint smoke detection framework to locate and recognize smoke from videos, in which a faster RCNN is used to generate suspected smoke region proposals and 3D CNN is used to extract temporal information. Although this method achieved better smoke detection performance compared with image-based methods, large computational cost limited its practical applications.

### B. Video Understanding

Although smoke detection from video is still a challenging task, great breakthrough has been made in video understanding [10]-[14][12]-[16].

Long-term Recurrent Convolutional Networks (LRCNs) [10] combined Convolutional Neural Network (CNN) and Recurrent Neural Network (RNN) for activity recognition and video description. 2D CNN was applied to process individual frames and output representation of image features for stack of LSTMs. Zhou *et al.* [11] proposed Temporal Relation Network (TRN) to learn and reason about temporal dependencies between video frames at multiple time scales. TRNs aims to describe the temporal relations between the spatial features extracted by 2D CNN. Temporal Shift Module (TSM) [12] was designed to solve video understanding from another side, which shifts part of the channels along the temporal dimension, both forward and backward, to exchange information among neighboring frames. Above methods explored to extract spatial feature through 2D

CNN, and then relied on different temporal model methods in the middle or output layer of 2D CNN.

Another research direction of video understanding is 3D CNN methods and its (2+1)D CNN variants. Tran *et al.* [15] proposed to use 3D CNN to model appearance and motion information simultaneously for videos. Carreira *et al.* [16] introduced a Two-Stream Inflated 3D ConvNet (I3D) that is based on 2D ConvNet inflation for video, and proved that, after pre-training on a large video dataset, performance of I3D models can be considerably improved. In this work, I3D models and its variants are considered as baseline methods on video smoke detection dataset.

Here, we explore to solve the smoke detection using video classification methods. The goal of this work is to explore whether the spatial and temporal branches in the smoke detection model could work together to improve the modeling ability of the model for smoke characteristics. Our key motivation of STCNet is that: a) subtle smoke motion feature may be highlighted by a simple and efficient residual frame calculation method; b) multi-scale spatio-temporal dual pyramid may be able to better integrate spatial and temporal information in the middle layers of two-path network; c) a coherent two-path network could take both temporal and spatial characteristics into account in the smoke detection process.

## III. SPATIO-TEMPORAL CROSS NETWORKS

In this section, we give a detailed description of the proposed video smoke detection method. The intuition behind the proposed methods is introduced firstly. Second, we present the architectures of Spatio-Temporal Cross Network. After that, the multi-level spatio-temporal representation structure details are reported, which are very important to decomposition of spatial and temporal components for smoke videos. Finally, we describe the spatio-temporal feature cross operation.

### A. Intuition

For video smoke detection task, it is a natural idea to apply existing video understanding methods directly to detect smoke from videoes. However, experiments suggest that general video understanding frameworks are not good at dealing with light smoke objects. These methods seem to pay more attention to obvious motion information. For the RISE dataset studied in this paper, industrial smoke objects usually have no obvious contours and texture features. Moreover, some samples are hard to recognize from single image, even for human eyes. In addition, due to the lack of effective smoke feature descriptors, it is difficult to generate the smoke optical flow representation. Detailed experimental results will be shown in the Section IV-C.

Inspired by two-stream methods, the proposed STCNet uses residual frames to focus on subtle temporal features of moving smoke areas. The intuition behind the STCNet is that spatial and temporal branches may guide each other and recognize smoke objects cooperatively. We assume that residual frames and RGB frames could focus on motion information and texture semantic information respectively. The cross fusion of these information may be helpful for detection model to make the final prediction.

### B. Overview of Methodology

With the assumption that the magnitudes of residual frames usually correlate with the smoke motion regions, we devise the Spatio-Temporal Cross Network (STCNet) as shown in Figure 3. For smoke video detection task, an input video is split into N subsections of same size, one RGB frame is sampled in each subsection (N=8 in this work). By jointly processing a small number of frames sampled from a whole video, the most relevant information of smoke objects can be captured. At the same time, processing fewer frames can reduce the inference time of model.

The proposed method is a two-path architecture, which consists of a spatial path using a CNN to extract smoke texture features and a temporal path using a same CNN (non-weight sharing) to calculate smoke motion features. Let $Frame_i \in \mathbb{R}^{C \times T \times H \times W}$, $ResFrame_i \in \mathbb{R}^{C \times T \times H \times W}$ be the $i_{th}$ RGB frames and $i_{th}$ residual frames in the spatial and temporal network respectively, with $C$, $T$, $H$, $W$ being the number of channels, temporal length, height and width of the image. The channel number $C$ is 3 for both RGB frames and residual frames. The RGB frames and residual frames are processed by spatial and temporal network respectively.

We assume that the two paths could guide each other. For example, the spatial path can easily filter out the obvious distractions such as swaying trees, while the temporal path can highlight the subtle traces of smoke. If they could guide each other, it would be helpful to improve the performance of the model. Therefore, we perform multi-scale feature cross fusion between the two branches.

#### 1) Spatial path

The backbone design of neural network is very important in the whole framework design. In recent years, many famous network structures have been designed for image classification, such as ResNet [22], ResNext [23], and so on. Here, following these successful network structures in image classification, we adapt them to design our Spatio-Temporal Cross Network for smoke detection.

For spatial path, stacked frames for one batch is $C \times T \times H \times W$. However, it should be noted that there is no temporal interaction between the stacked frames in the inference of spatial branch. The design goal of space path is to focus on the texture and appearance of smoke regions. Therefore, each input frame is processed as an independent individual.

Multi-scale spatial feature pyramid is extracted by spatial CNN backbone. The detail of spatio-temporal feature pyramid will be reported in the following.

#### 2) Residual frames

For the industrial smoke detection task, one of the most challenging problems is that smoke features are not as intuitive as general object recognition. Therefore, we hope to obtain residual frames by subtracting adjacent frames to highlight the changing regions between frames. In order to reserve color and long-term dependence information, stacked residual frames of RGB channel are set as temporal path input.

Assuming the $i_{th}$ RGB frame is formulated as $Frame_i$, the $i_{th}$ residual frame can be define as:

$$ResFrame_i = \begin{cases} \alpha * |Frame_i - Frame_{i+1}|, & if < \beta \\ \beta, & if \geq \beta \end{cases}$$

where the $\alpha$ is an expanding coefficient, which highlights the frame differences. Then, the maximum residual frame pixel value is limited with the parameter $\beta$ ( $\beta = 255$ in experiments), its function is to prevent numerical overflow. Some RGB and corresponding frames are shown in Figure 1. The first row shows the normal RGB images, in which the red arrow indicates the approximate location of smoke. The smoke area in residual frames usually contains light cyan components. Compared with RGB frames, the information in residual images is relatively sparse, and they mainly focus on moving objects.

The subtle smoke feature is enhanced by the expanding coefficient $\alpha$, and then the maximum pixel value is limited to 255 by parameter $\beta$. These operations not only highlight the characteristics of smoke, but also suppress the interference such as steam.

For each frame, we only need to calculate the difference with the next frame. The computational cost can even be ignored when compared with the convolutional neural network latency or optical flow calculation.

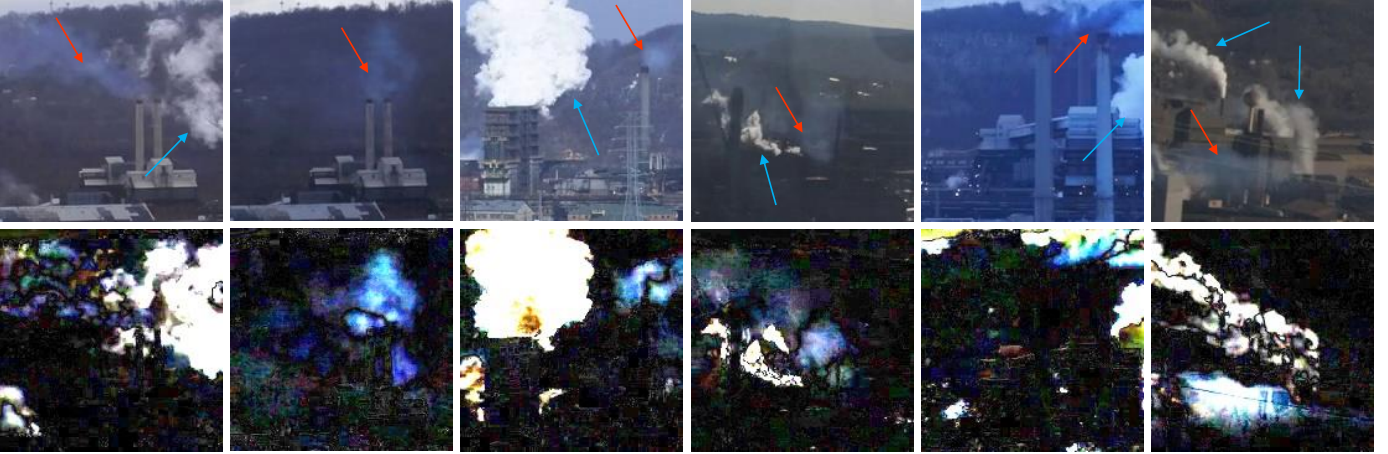

**Figure 1.** Input RGB frames (the top row) in RISE dataset and corresponding residual frames (the bottom row). For clarity, red and blue arrows are used to mark smoke and steam (non smoke) respectively in the input images.

### 3) Temporal path

In parallel to the spatial path, temporal path is another CNN model designed to capture smoke motion features. After obtaining the residual frames, convolutional neural network can extract motion features from the frame differences. In order to reduce the complexity of the network structure design, the architecture of temporal network is the same as that of spatial network. In order to capture different spatial and temporal information, the weights of spatial and temporal branches are different. The spatial path focuses on the smoke appearance and background information, while the temporal branch focuses on the moving region.

Our spatio-temporal cross network architecture is generic, and it can be instantiated with different backbones, such as Se-ResNext and MobilNetv2. Generally, the complex backbone can obtain better recognition performance, while the advantage of lightweight model lies in the computational speed. An STCNet example using SE-ResNeXt-50 as backbone is specified in Table I.

### C. Multi-level Structure

A deep CNN model generates feature maps layer by layer, and with pooling layers the feature maps have different sizes and depths. The high-resolution in-network feature maps have low-level features which contain detailed spatial information. However, the low-resolution feature maps have high-level semantic features which have stronger representational capacity for object recognition. The Feature Pyramid Network (FPN) [45] is a common technique in object detection and image classification task to improve the feature representation ability of the CNN.

TABLE I
AN EXAMPLE INSTANTIATION OF THE STCNET.

| Stage | Spatial path | Temporal path | Output sizes (T×C×W×H) |
|---|---|---|---|
| data | - | - | Spatial: T×3×224² Temporal: T×3×224² |
| conv1 | 1×7², 64 stride 2² | 1×7², 64 stride 2² | Spatial: T×3×224² Temporal: T×3×224² |
| pool1 | 1×3², max stride 2² | 1×3², max stride 2² | Spatial: T×64×56² Temporal: T×64×56² |
| res1 | $\begin{bmatrix} 1\times1^2, 128 \\ 1\times3^2, 128 \\ 1\times1^2, 256 \end{bmatrix}$ ×3 | $\begin{bmatrix} 1\times1^2, 128 \\ 1\times3^2, 128 \\ 1\times1^2, 256 \end{bmatrix}$ ×3 | Spatial: T×256×56² Temporal: T×256×56² |
| res2 | $\begin{bmatrix} 1\times1^2, 256 \\ 1\times3^2, 256 \\ 1\times1^2, 512 \end{bmatrix}$ ×4 | $\begin{bmatrix} 1\times1^2, 256 \\ 1\times3^2, 256 \\ 1\times1^2, 512 \end{bmatrix}$ ×4 | Spatial: T×512×28² Temporal: T×512×28² |
| res3 | $\begin{bmatrix} 1\times1^2, 512 \\ 1\times3^2, 512 \\ 1\times1^2, 1024 \end{bmatrix}$ ×6 | $\begin{bmatrix} 1\times1^2, 512 \\ 1\times3^2, 512 \\ 1\times1^2, 1024 \end{bmatrix}$ ×6 | Spatial: T×1024×14² Temporal: T×1024×14² |
| res4 | $\begin{bmatrix} 1\times1^2, 1024 \\ 1\times3^2, 1024 \\ 1\times1^2, 2048 \end{bmatrix}$ ×3 | $\begin{bmatrix} 1\times1^2, 1024 \\ 1\times3^2, 1024 \\ 1\times1^2, 2048 \end{bmatrix}$ ×3 | Spatial: T×2048×7² Temporal: T×2048×7² |
| fuse | conv, 1×1², 256 | | T×256×7² |
| cls | conv, 1×1², 256 adaptive average pool | | 1×256×1² |
| out | fully connected layer | | # classes |

In STCNet, in order to effectively fuse spatial and temporal features from multi-scales, a dual pyramid fusion structure of spatio-temporal features for dual path was designed. Our method takes T RGB and residual frames as input, and outputs spatial and temporal feature maps at several scales with a scaling step of 2. There are often many residual blocks generating same size feature maps and we combine these blocks into the same stage (as shown in Table I). For our multi-level structure, the output of the last layer of each residual block in spatial and temporal path is chosen to build the dual feature pyramid.

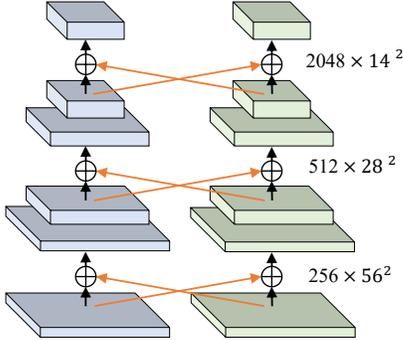

**Figure 2.** The network architecture of multi-scale spatio-temporal feature pyramid (SE-ResNext-50 as backbone).

Specifically, each residual stage of SE-ResNext-50 is denoted as {res1, res2, res3, res4}. Their output have stride of {4, 8, 16, 32} with respect to the input frame. The first three pairs of feature maps are selected to build the dual pyramid. The last 7×7 size feature map is fused for the final classification head prediction, rather than as the top layer of the pyramid.

In addition, we also design a dual pyramid variant, in which connections fuse from the temporal to the spatial path. The detail results will be reported in Section IV-D.

### D. Cross Operation

As shown in Figure 2, a spatio-temporal dual pyramid sample is constructed to improve the smoke feature recognition ability. In this structure, the feature maps of spatial and temporal path are summed with each other to participate in model inference. We assume that the dual pathway could guide each other and make progress together in this way. Specifically, the sum fusion operation of two feature maps can be define as:

$$SF_{i,j,c}^{sum} = TF_{i,j,c}^{sum} = SF_{i,j,c} + TF_{i,j,c}$$

where $1 \leq i \leq H$, $1 \leq j \leq W$, $1 \leq c \leq C$ and $SF, TF \in \mathbb{R}^{C \times H \times W}$. Since the feature maps to be fused come from the same location of neural networks with the same architecture. Therefore, the sum fusion is element-wise addition between two feature maps.

Although the summed features in two paths are same, the focus of two branches are different. In Section IV-E, we will show the activated regions on 7×7 size feature maps of two branches by Grad-CAM [46], which helps us to analyze whether the model working as expected.

In order to make the research more convincing, we discuss different ways of fusing methods between two pathway. Conv fusion is another common feature fusion method. We also design to change the orange arrow and addition operation into convolution fusion in Figure 2. However, two problems are exposed, one is the slow convergence speed, another is the reduction of accuracy. Finally, identity mapping is used in the orange arrows in this work.

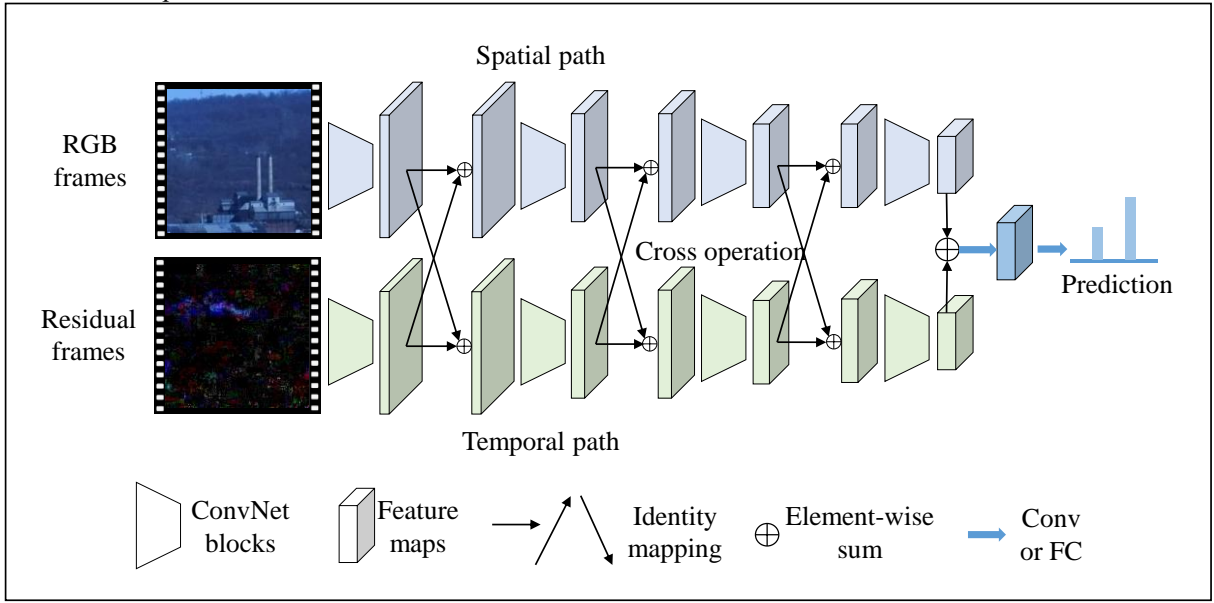

**Figure 3.** Framework of the proposed Feature Foreground Enhanced Network (FFENet) for smoke source prediction and detection.

## IV. EXPERIMENTS

In this section, we evaluate proposed methods on RISE industrial smoke emission dataset. Experiments were conducted on a personal computer with CPU of Intel Core i7-9700 and GPU of NVIDIA RTX2080TI using the Pytorch framework.

### A. RISE Dataset

The RISE video smoke dataset [29] is the first large-scale video dataset for recognizing industrial smoke emissions. RISE dataset contains 12,567 clips with 19 distinct views from cameras on three sites that monitored three different industrial facilities. The clips are from 30 days that spans four seasons in two years in the daytime. RISE is a challenging video classification dataset, as it covers various characteristics of smoke emissions, including opacity and color, under diverse weather (e.g., haze, fog, snow, cloud) and lighting conditions. Moreover, RISE involves distractions of various types of steam, which can be similar to smoke and challenging to distinguish.

### B. Implementation Details

Stochastic Gradient Descent (SGD) with a mini-batch size of 3 is used to optimize model weights. The weight decay is 0.0005 and the momentum is 0.9. We start from a learning rate of 0.001 and a pre-trained SE-ResNeXt-50 on ImageNet because of its balance between accuracy and efficiency. In addition, the proposed framework based on other backbone models has also been tested, such as MobileNetv2 [47].

### C. Comparisons on RISE

We compare the proposed method with primary methods on video understanding [10] [12] [14] [16] [35]. The comparison results are shown in Table II. Abbreviation ND and FP means no data augmentation and frame perturbation, respectively. Abbreviation TSM, LSTM, NL and TC means Temporal Shift module [12], Long Short-Term Memory layers [10], Non-Local module [14], and Timeception layers [35], respectively.

Results in Table II show that plain CNN model based on SE-ResNeXt-50 achieves a better performance than the original baseline methods, which show that a stronger backbone is helpful for smoke recognition. When adopting 8 frames as input, our STCNet gains 6.2% higher F-score than the RGB-I3D-TC method, which confirms the remarkable ability of STCNet for smoke temporal modeling. Data augmentation technology can further improve the performance of STCNet. In training stage, The data augmentation methods (horizontal flipping, random resizing and cropping, perspective transformation, area erasing, and color jittering) same as in [29] is used by default.

Moreover, the parameters, FLOPs, latency and throughput characterization of each method are also reported in Table III. Generally, the proposed STCNet can achieve 42.57 videos per second processing speed. Although the STCNet model of SE-ResNext backbone has no obvious advantage in processing speed, we also designed the STCNet based on MobileNetv2, which achieves a much faster computing speed than other methods: 109.7 videos per second, as shown in Table III.

TABLE II
F-SCORES FOR COMPARING DIFFERENT METHODS ON RISE DATASET.

| Model | $S_0$ | $S_1$ | $S_2$ | $S_3$ | $S_4$ | $S_5$ | Average |
|---|---|---|---|---|---|---|---|
| Flow-SVM | .42 | .59 | .47 | .63 | .52 | .47 | .517 |
| Flow-I3D | .55 | .58 | .51 | .68 | .65 | .50 | .578 |
| RGB-SVM | .57 | .70 | .67 | .67 | .57 | .53 | .618 |
| RGB-I3D [16] | .80 | .84 | .82 | .87 | .82 | .75 | .817 |
| RGB-I3D-ND [16] | .76 | .79 | .81 | .86 | .76 | .68 | .777 |
| RGB-I3D-FP [16] | .76 | .81 | .82 | .87 | .81 | .71 | .797 |
| RGB-I3D-TSM [12] | .81 | .84 | .82 | .87 | .80 | .74 | .813 |
| RGB-I3D-LSTM [10] | .80 | .84 | .82 | .85 | .83 | .74 | .813 |
| RGB-I3D-NL [14] | .81 | .84 | .83 | .87 | .81 | .74 | .817 |
| RGB-I3D-TC [35] | .81 | .84 | .84 | .87 | .81 | .77 | .823 |
| Plain SE-Resnext | .83 | .82 | .84 | .85 | .78 | .83 | .826 |
| **STCNet(MobileNetv2)** | **.86** | **.88** | **.87** | **.89** | **.84** | **.86** | **.868** |
| **STCNet(SE-ResNext)** | **.88** | **.89** | **.90** | **.90** | **.86** | **.88** | **.885** |

TABLE III
COMPARISON WITH OTHER METHODS ON RISE DATASET.

| Model | Backbone | Params | Flops | Latency | Throughput | Average |
|---|---|---|---|---|---|---|
| RGB-I3D [16] | Inception I3D | 12.3M | 62.7G | 30.56ms | 32.71vid/s | .817 |
| RGB-I3D-TSM [12] | Inception I3D | 12.3M | 62.7G | 31.85ms | 31.40vid/s | .813 |
| RGB-I3D-LSTM [10] | Inception I3D | 38.0M | 62.9G | 31.01ms | 32.25vid/s | .813 |
| RGB-I3D-NL [14] | Inception I3D | 12.3M | 62.7G | 30.32ms | 32.98vid/s | .817 |
| RGB-I3D-TC [35] | Inception I3D | 12.3M | 62.7G | 30.41ms | 32.88vid/s | .823 |
| Plain SE-Resnext | SE-ResNeXt-50 | 26.6M | 34.4G | 22.10ms | 45.25vid/s | .826 |
| **STCNet** (Proposed) | Mobilenetv2 | **3.7M** | **2.4G** | **9.12ms** | **109.7vid/s** | .868 |
| **STCNet** (Proposed) | SE-ResNeXt-50 | 27.2M | 34.6G | 23.49ms | 42.57vid/s | **.885** |

### D. Comparison with other variants

At the beginning of architecture design, we also try some test schemes to gradually explore the optimal detection architecture.

Here, three STCNet variants are designed for comparison, which are denoted as STCNet-A (Figure 4), STCNet-B (Figure 5) and STCNet-C (Figure 6). Among them, STCNet-A is a typical two-stream network, which sums the output feature maps of two pathway to predict probability. The difference is that there is no optical flow input, but the residual frames calculated in the way of Section III-B-2). STCNet-B is equipped with unidirectional feature fusion method, which only fuses temporal to spatial feature. The aim of designing STCNet-B is to verify the efficiency of fusion operation from spatial to temporal path. In addition, STCNet-C performs the spatio-temporal feature fusion only after the first residual block without our scale feature fusion operation.

Three variant models are trained and tested on RISE dataset. The same hyperparameters are adopted for three models. The results are reported in Table IV.

With the multiscale fusion architecture, the STCNet increases by 1.3% compared with STCNet-A. After adding multi-scale fusion operation from temporal path to spatial path in STCNet-A, the F-score of STCNet-B is improved to 0.882. But compared with the STCNet of bidirectional fusion, it is 0.3% behind. Finally, the F-score of STCNet-C is reduced by 0.7% using only shallow feature fusion. These ablation experiments prove the effectiveness of the multi-scale spatio-temporal feature fusion operation in STCNet.

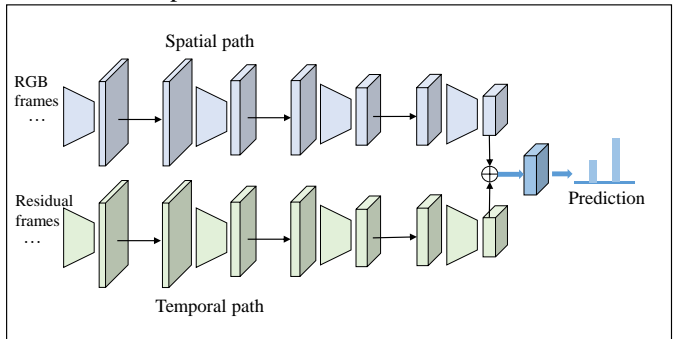

**Figure 4.** Framework of the STCNet-A (similar to Two-stream network [39]).

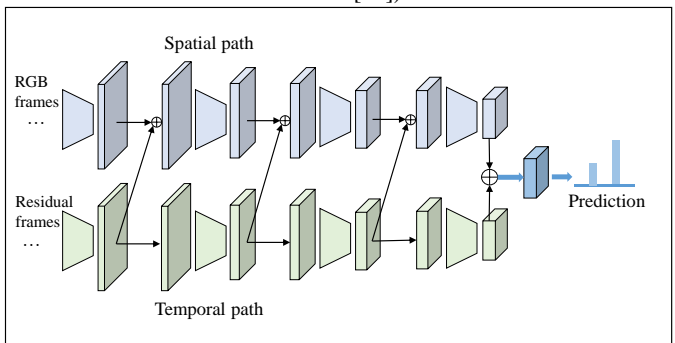

**Figure 5.** Framework of the STCNet-B, in which internal connections fuse from the temporal to the spatial path.

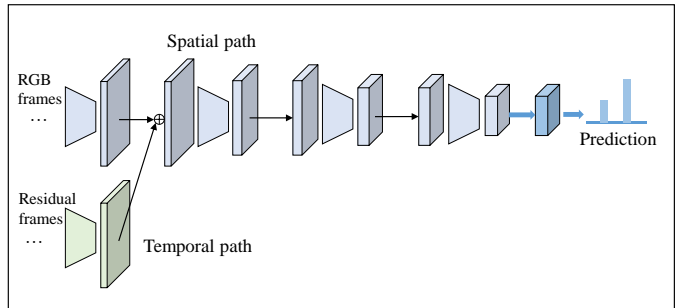

**Figure 6.** Framework of the STCNet-C, which performs single scale feature fusion near the residual frame input.

TABLE IV
F-SCORES FOR COMPARING DIFFERENT VARIANTS ON RISE DATASET.

| Model | $S_0$ | $S_1$ | $S_2$ | $S_3$ | $S_4$ | $S_5$ | Average |
|---|---|---|---|---|---|---|---|
| **STCNet-A** | .87 | .88 | .89 | .89 | .84 | .86 | .872 |
| **STCNet-B** | .88 | .89 | .90 | .90 | .85 | .87 | .882 |
| **STCNet-C** | .88 | .89 | .90 | .88 | .85 | .86 | .877 |
| **STCNet** | **.88** | **.89** | **.90** | **.90** | **.86** | **.88** | **.885** |

### E. Visualization of two path

In order to diagnose the attention of two pathway in STCNet, we apply the Gradient-weighted Class Activation Mapping (Grad-CAM [46]) to visualize the active regions in input frames. The 7×7 output of the last residual blocks of two branches is visually analyzed, as shown in Figure 7. For most of the input frames, the proposed method can focus on the smoke region in two path, while ignoring the interference of steam (non smoke). For example, in the second column, even if human eyes are difficult to find smoke, the temporal path can well focus on the region of smoke movement. It is difficult to detect these subtle smoke information only by spatial feature extraction.

These cases with obvious steam interference confirm the excellent detection ability of proposed STCNet.

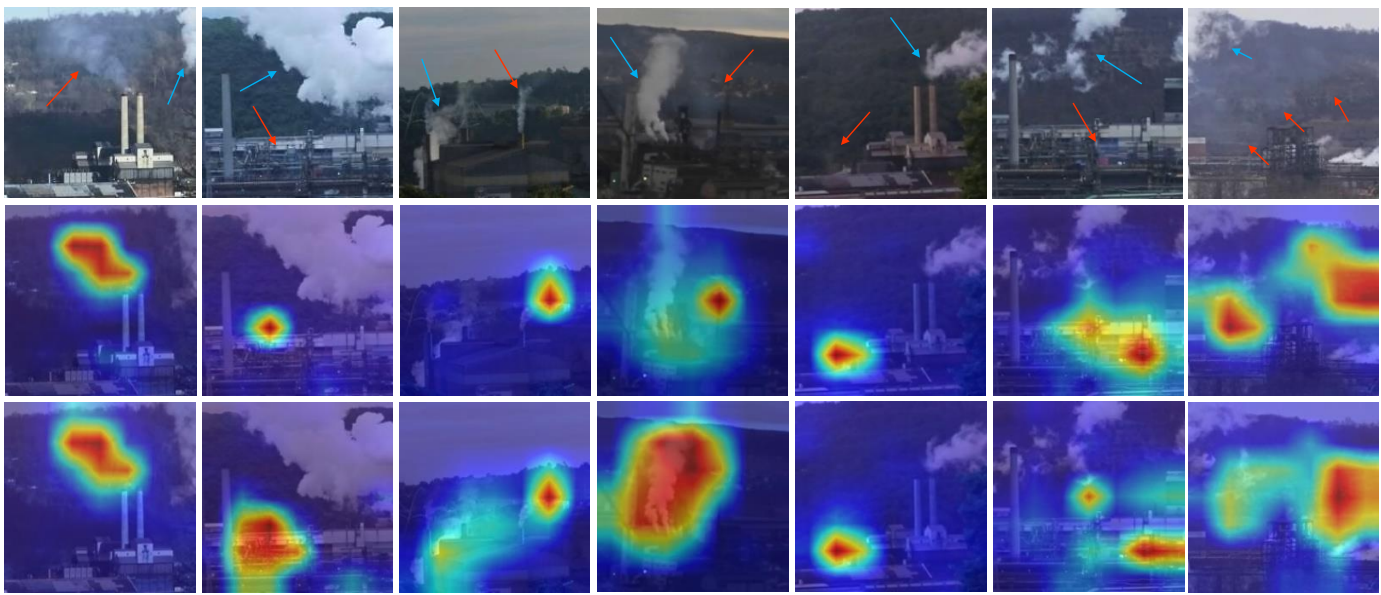

**Figure 7.** Grad-CAM visualization for spatial and temporal pathway. For clarity, red and blue arrows are used to mark smoke and steam (non smoke) respectively in input images. The second row is visualization for spatial path, and the last row is for temporal pathway.

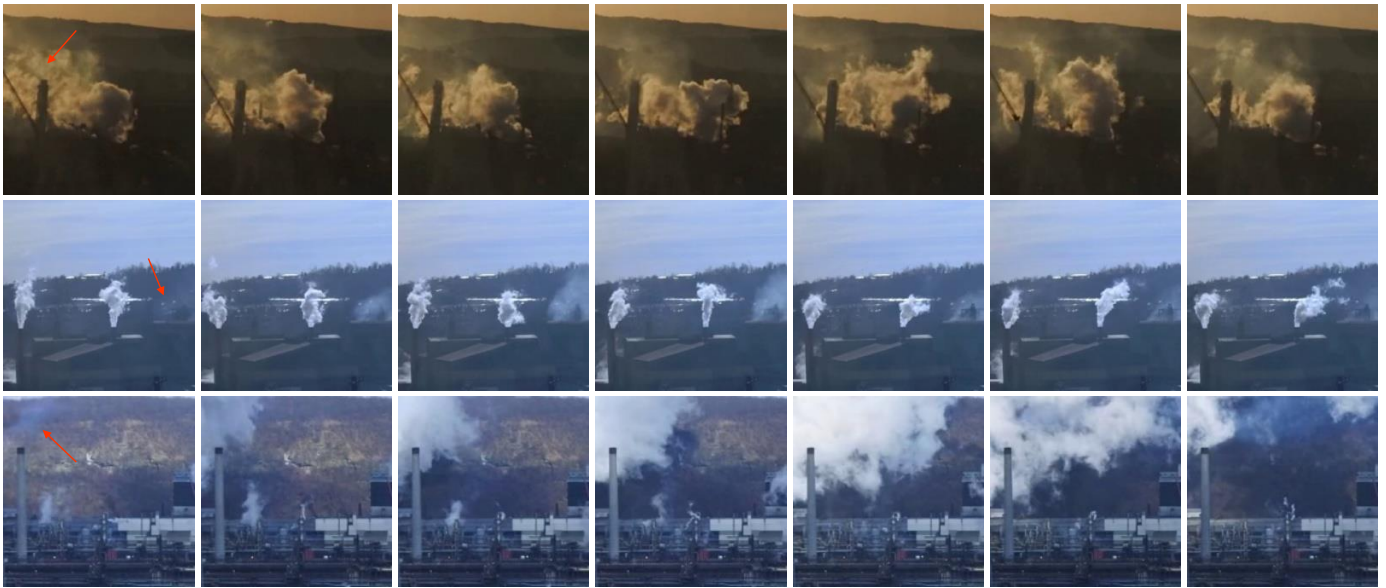

**Figure 8.** False positive cases in the testing set. The undetected smoke regions is marked by red arrows.

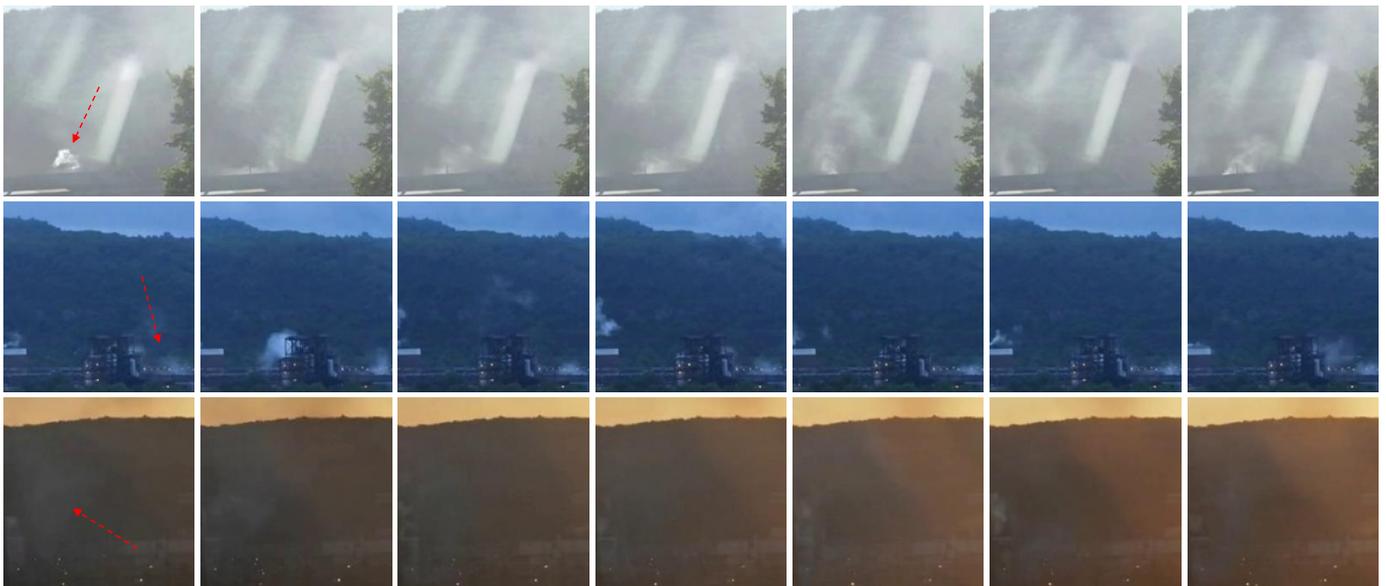

**Figure 9.** False negative cases in the testing set. Areas that may cause false positives are marked with red dotted arrows.

Finally, we show some false positive and false negative cases in Figure 8 and Figure 9, respectively. We analyze the deficiencies of the proposed method through these cases.

In Figure 8, red arrows are used to mark smoke regions which cannot be detected. The smoke in the first and third rows is almost completely obscured by steam. Smoke in second and third rows have a small amount for a short duration. These may be the reasons why the model failed to detect smoke.

Some false positive cases are shown in Figure 9. We used red dotted arrows to mark the regions that cause false positives. In the first two rows, light steam leads to false detection results. The third cause of false alarm is the large moving fog, which may mislead the model to make wrong predictions. In fact, small areas and short duration steam or smoke object detection is still a challenging task.

## CONCLUSION

In this work, a novel Spatio-Temporal Cross Network (STCNet) was designed and verified for video smoke detection task. Inspired by two-stream methods, spatio-temporal dual pyramid architecture was proposed to focus on efficient spatio-temporal information fusion. Extensive experiment results on challenging smoke dataset demonstrated the remarkable ability of STCNet in smoke detection.

However, the task of industrial smoke detection is far from being solved, and many challenges still remain, such as the classification between smoke and steam. For further research, we will study on the fine-grained category classification between smoke and steam in videos.